\documentclass[letterpaper, 10 pt, conference]{ieeeconf}  % Comment this line out if you need a4paper

\IEEEoverridecommandlockouts                              % This command is only needed if 
\title{\LARGE \bf A 3D Printing Hexacopter: Design and Demonstration}

\author{Alexander Nettekoven$^{1}$ and Ufuk Topcu$^{2}$% <-this % stops a space
\thanks{$^{1}$Alexander Nettekoven is with the Walker Department of Mechanical Engineering at the University of Texas at Austin, 204 E. Dean Keeton Street, Austin, Texas 78712, USA
        {\tt\small nettekoven@utexas.edu}}%
\thanks{$^{2}$Ufuk Topcu is with the Faculty of the Department of Aerospace Engineering and Engineering Mechanics at the University of Texas at Austin, 2617 Wichita Street, Austin, Texas 78712-1221, USA
	{\tt\small utopcu@utexas.edu}}%
}

\usepackage{graphicx}
\usepackage{color, soul}
\usepackage{gensymb}
\usepackage{amsmath}
\usepackage{pgfplots}
\usepackage{subfigure}

\begin{document}

\maketitle
\thispagestyle{empty}
\pagestyle{empty}

%%%%%%%%%%%%%%%%%%%%%%%%%%%%%%%%%%%%%%%%%%%%%%%%%%%%%%%%%%%%%%%%%%%%%%%%%%%%%%%%
\begin{abstract}
	
3D printing using robots has garnered significant interest in manufacturing and construction in recent years. A robot's versatility paired with the design freedom of 3D printing offers promising opportunities for how parts and structures are built in the future. However, 3D printed objects are still limited in size and location due to a lack of vertical mobility of ground robots. These limitations severely restrict the potential of the 3D printing process. To overcome these limitations, we develop a hexacopter testbed that can print via fused deposition modeling during flight. We discuss the design of this testbed and develop a simple control strategy for initial print tests. By successfully performing these initial print tests, we demonstrate the feasibility of this approach and lay the groundwork for printing 3D parts and structures with drones.
	 
\end{abstract}

%%%%%%%%%%%%%%%%%%%%%%%%%%%%%%%%%%%%%%%%%%%%%%%%%%%%%%%%%%%%%%%%%%%%%%%%%%%%%%%%

\section{INTRODUCTION}

3D printing has fundamentally changed the way 3D parts and structures are designed and produced~\cite{am_review}. The design freedom and the convenience of printing 3D objects directly from computer-aided design are just a few advantages that have been continuously leading to fundamental changes in established processes of these industries, e.g., the printing of houses~\cite{am_review, concrete_printing_examples2, D3D_review1, D3D_review3}. Recent work has focused on combining robotics with 3D printing as the versatility of robots can improve the printing of medium to large-scale objects~\cite{am_review, concrete_printing_examples2, D3D_review1, D3D_review3, D3D_review4}. However, a lack of vertical mobility and sometimes horizontal mobility for ground robots still significantly limits the 3D printing of these objects~\cite{concrete_printing_examples, concrete_printing_examples2, D3D_review4}.

At the same time, rapid advancements in control and autonomy of drones are driving the exploration of new drone applications in manufacturing and construction that go far beyond already established surveillance and mapping applications~\cite{drones_in_manufacturing, aeroarms}. Prominent examples are pick-and-place tasks similar to robotic arms and surface inspections that require close proximity to objects of interest during flight~\cite{drones_in_manufacturing, aeroarms, ETH_pick_place}. Unlike ground robots and manufacturing machines, drones have the advantage of being unrestricted in 3D space and are more capable of reaching remote locations. A merger of 3D printing and drone technology has the potential to overcome these limitations and fundamentally enhance the way 3D objects are built for the manufacturing and construction industries~\cite{drones_in_manufacturing, am_with_drones_review}.

In this paper, we develop a hexacopter testbed that can print simple contours on a build surface while flying. By successfully performing initial print tests, we lay the groundwork to print entire 3D parts and structures in future work. Successful printing may not only provide insights for 3D printing applications with drones but also for other manufacturing applications with drones that require surface contact or close to ground operations. 

\begin{figure}
	\centering
	\includegraphics[width=0.9\linewidth, trim={35cm 15cm 5cm 0cm}, clip=true]{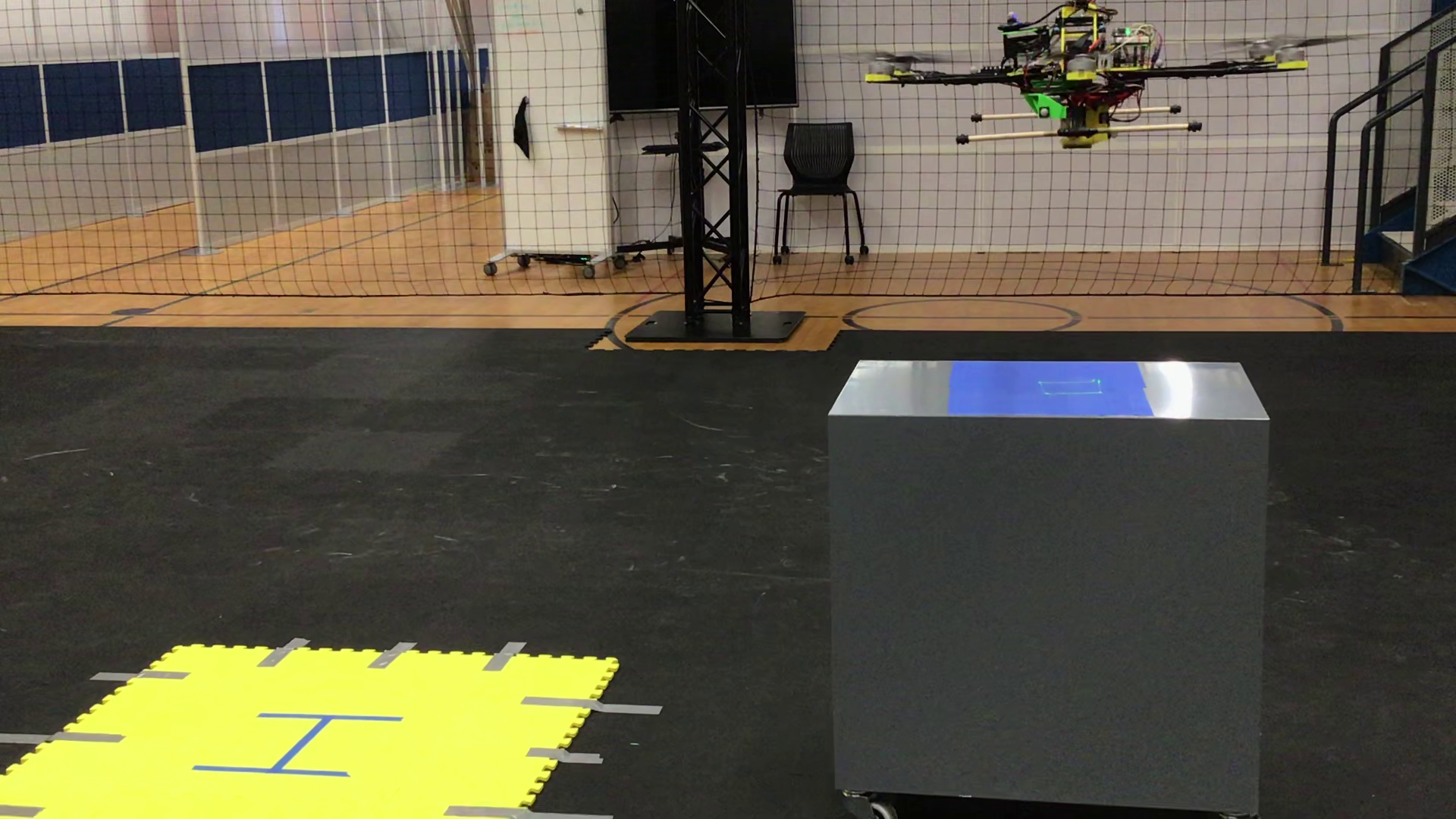}
	\caption{The 3D printing hexacopter just finished printing a square contour on top of a gray box. A video of this print test can be found at \it{https://youtu.be/tEooDpE2TyE}.}
	\label{fig:printingimg}
\end{figure}

We structure the remainder of the paper as follows. In Section~\ref{sec2}, we provide a brief overview of related robotics work in manufacturing and construction. In Section~\ref{sec3}, we detail the design of the 3D printing hexacopter. Section~\ref{sec4} details the control strategy for achieving the desired printing task. In Section~\ref{sec5}, we present and discuss the first successful prints with the 3D printing hexacopter. Finally, we provide our conclusions and discuss the next steps in Section~\ref{sec6}.

\section{RELATED WORK}\label{sec2}

Numerous robotic testbeds have demonstrated the successful combination of 3D printing with robots; a selection of these systems and their printing capabilities can be found in recent surveys for manufacturing and construction~\cite{D3D_review1, D3D_review3, D3D_review4, concrete_printing_examples, concrete_printing_examples2}.  Notable examples include the multi-robot printing system by~\cite{multi_printing_system} or the 3D printing system by~\cite{robotic_arm_printing}. Another example is a climbing 3D printer system that can scale its printed structures to continue printing in the vertical direction~\cite{Koala3D}. These systems provide promising solutions for printing large 3D objects. However, these systems are still restricted in 3D space and depend on the structural integrity of the 3D printed objects.

To overcome these limitations, initial work by~\cite{flying_robot_3D} investigated the combination of a quadcopter with 3D printing by using a syringe-like depositing apparatus for different polymer materials. These materials were in a highly viscous liquid form and were deposited by the quadcopter from a considerable distance to the surface. The deposited material covered large areas of the surface, including areas beyond the desired printing location. Due to the large spread of the deposited material, this approach was not able to print 3D objects with distinct shapes. Recent work by~\cite{recent_3D_printing_drone} investigated different materials that could be used for a similar depositing setup as~\cite{flying_robot_3D}. Though the authors identify suitable materials for different applications, the depositing apparatus still suffers from the same limitations as~\cite{flying_robot_3D}.

\section{DRONE DESIGN}\label{sec3}

The goal of the hexacopter testbed is to deposit material with high positional accuracy. In other words, the hexacopter should be able to move to a desired location and deposit material precisely at that location. Since hexacopters have less positional accuracy than robotic arms or other 3D printing machines and also have a considerable amount of vibrations, the hexacopter design plays a critical role in achieving this goal. For designing a 3D printing hexacopter, we first discuss which 3D printing process is used and why. Based on this 3D printing process, we derive several design objectives that guide the design of the different submodules.

\subsection{Selection of 3D printing process}

Fused deposition modeling (FDM) is the most popular 3D printing technology and is commonly used in desktop 3D printers~\cite{wohlers_report}. FDM offers several advantages, such as affordability, wide material choice, and the simplicity of the technology. The main components of the FDM printer consist of a hotend that heats the desired printing material, an extruder that pushes the material out of the extruder nozzle, and a mechanism that moves the FDM printer along a desired surface. Furthermore, the FDM process uses a thinner filament than concrete printing, predominantly used in related work. A thinner filament allows for finer printing quality and thus, a potentially better analysis of the movement mechanism of the 3D printer. In addition to these advantages, the FDM process is similar to other nozzle-based 3D printing processes and even non-3D printing processes that have a distinct tool tip similar to a nozzle. These similarities could give a testbed the potential to be used for other applications if desired and can make the 3D printing results relevant for applications beyond 3D printing. Due to these advantages, the FDM technology is picked for the 3D printing hexacopter. We use polylactic acid (PLA) as the printing material due to its ease-of-use for printing. 

\subsection{Design objectives}

Based on the selected 3D printing process, we derive several design objectives. First, the nozzle needs to be close to the build surface for optimal extrusion and adhesion to the build surface. If the hexacopter is too far from the build surface, the extruded material will be less likely to deposit at the desired location on the surface. This larger distance could adversely affect the printing accuracy or even cause the deposited material to fly away due to the turbulent air from the hexacopter. In FDM printers, this distance is usually less than a millimeter. 

Second, the extruder nozzle should not touch the surface as the surface might damage the nozzle and vice versa. Ideally, the distance between the nozzle and build surface stays constant to ensure uniform printing. However, since the hexacopter is flying close to the build surface, turbulent air from the build surface can pose significant challenges to keeping a tight, constant distance to the build surface. 

Third, the FDM hotend uses high temperatures to melt the printing material. Proper insulation of the hotend is essential to prevent the hotend from losing too much heat from the circulating air and to prevent the surrounding hexacopter parts from possible damage. Finally, to minimize the hexacopter's total weight, any parts added should be lightweight while still being sturdy enough to minimize vibrations during flight.

\subsection{Flight platform}

Several commercial flight platforms already exist to carry larger payloads, such as the equipment necessary for 3D printing. However, the integration of the mechatronic equipment and the desire to have full access to the underlying control algorithms make building a custom-built flight platform for the testbed a better option. 

\begin{figure}
	\centering
	\includegraphics[width=0.8\linewidth]{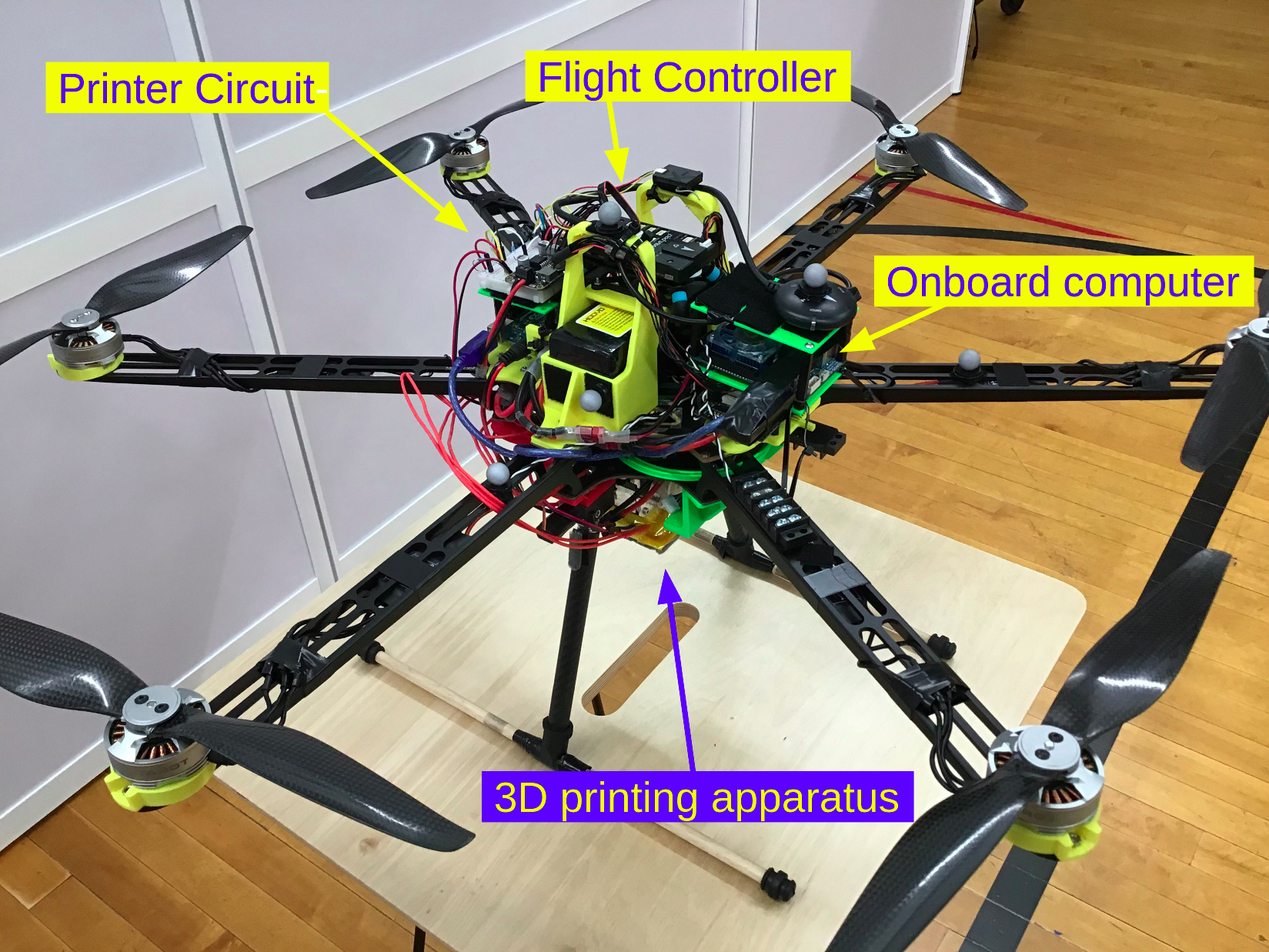}
	\caption{Top view of the 3D printing hexacopter. The main electronic components of the testbed are labelled.}
	\label{fig:overview}
\end{figure}

The core of the 3D printing hexacopter is taken from a DJI F550 ready-to-fly kit and consists of a top and bottom platform. The core is at the geometric center of the hexacopter and carries the flight electronics, the 3D printing equipment, and the batteries. Attached to the core are six custom arms that have been extended to accommodate larger motors than the standard F550 motors. The motors are mounted on 3D printed mounts that have a dihedral angle of eight percent. This dihedral angle increases the hexacopter's stability and significantly reduces vibrations caused by complex airflow interactions between the propellers and the ground, also called ground effects~\cite{dihedral}. Fig.~\ref{fig:overview} shows an overview of the testbed. 

The flight electronics consist of a Pixhawk 4 flight controller loaded with the open-source software PX4. The flight controller is connected to an onboard ODROID-XU4 computer that uses ROS and Wifi to communicate with an external laptop. The onboard computer also feeds position and velocity measurements from a \mbox{VICON} motion capture system to the flight controller. Section~\ref{sec4} provides details on the control of the hexacopter.

\subsection{3D printing apparatus}

The 3D printing apparatus consists of the extruder, the hotend, the printing material, the circuitry powering and controlling the printing process, and the nozzle guard. We describe the different components in the next paragraphs.

For the extruder and hotend, we use an E3D Titan Aero print head that combines both components in a compact form. The hotend of this print head is capable of reaching temperatures up to $285~\degree C$. We use a 1 mm diameter for the hotend's nozzle and attach the print head to the bottom of the flight platform with 3D printed connectors. These connectors are designed with generative design to find a sturdy solution that can withstand potential ground impacts during testing while also being light enough to reduce the hexacopter's weight. 

\begin{figure}
	\centering
	\includegraphics[width=0.8\linewidth]{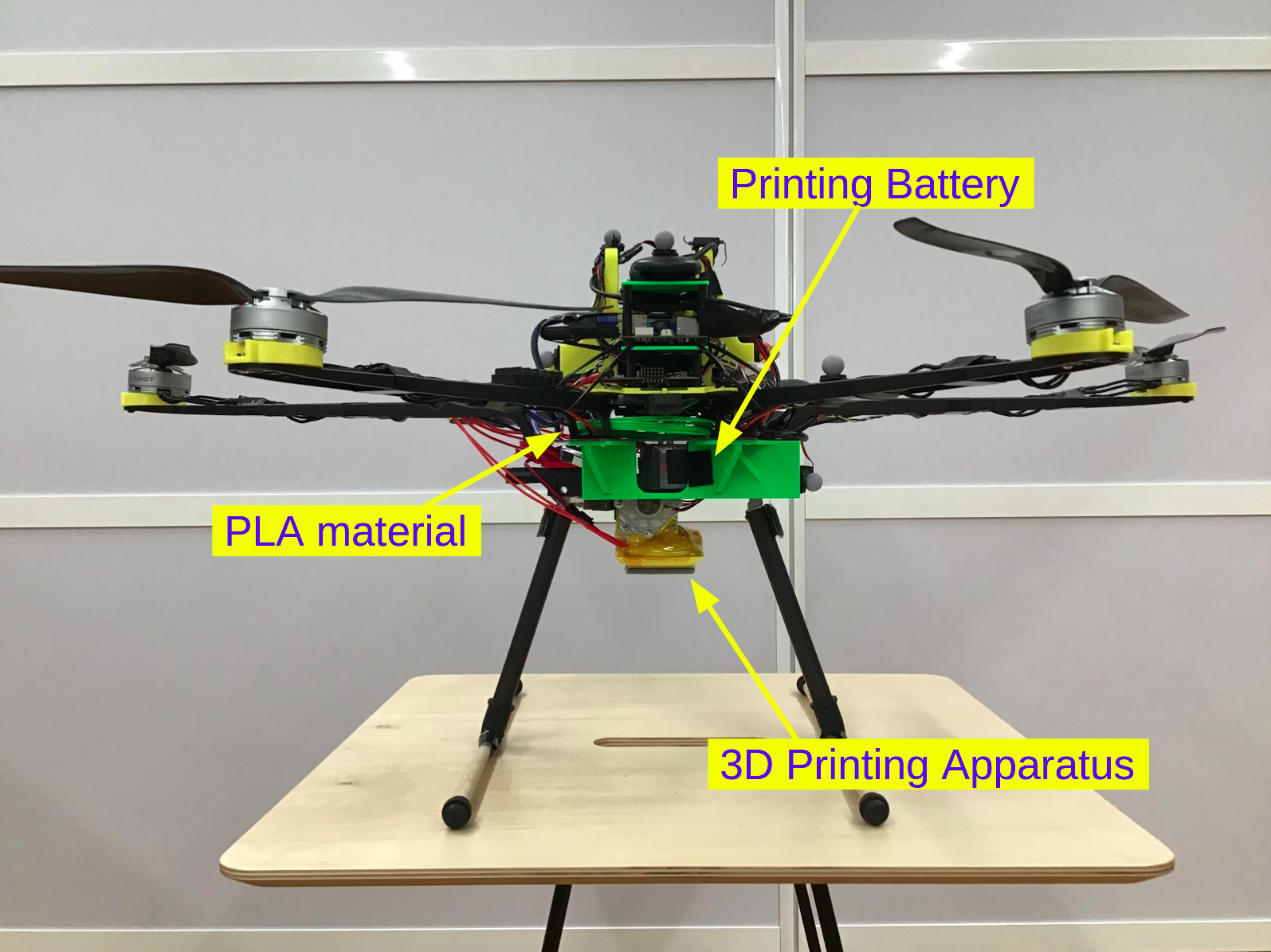}
	\caption{Front view of the 3D printing hexacopter. The main components for 3D printing are labelled.}
	\label{fig:frontView}
\end{figure}

For storing the 3D printing material, we use the space between the top and bottom platforms of the hexacopter (see Fig.~\ref{fig:frontView}). We feed the printing material through a hole in the bottom platform to the print head. The printing circuit that controls the 3D printing equipment has its own battery, which is located underneath the hexacopter next to the print head. Controlling the print head only requires an Arduino Uno as well as several basic electronic components, such as transistors and stepper drivers.

\begin{figure}
	\centering
	\subfigure[3D printing apparatus]
	{
		\includegraphics[height=0.3\linewidth, trim={20cm 40cm 20cm 40cm}, clip=true]{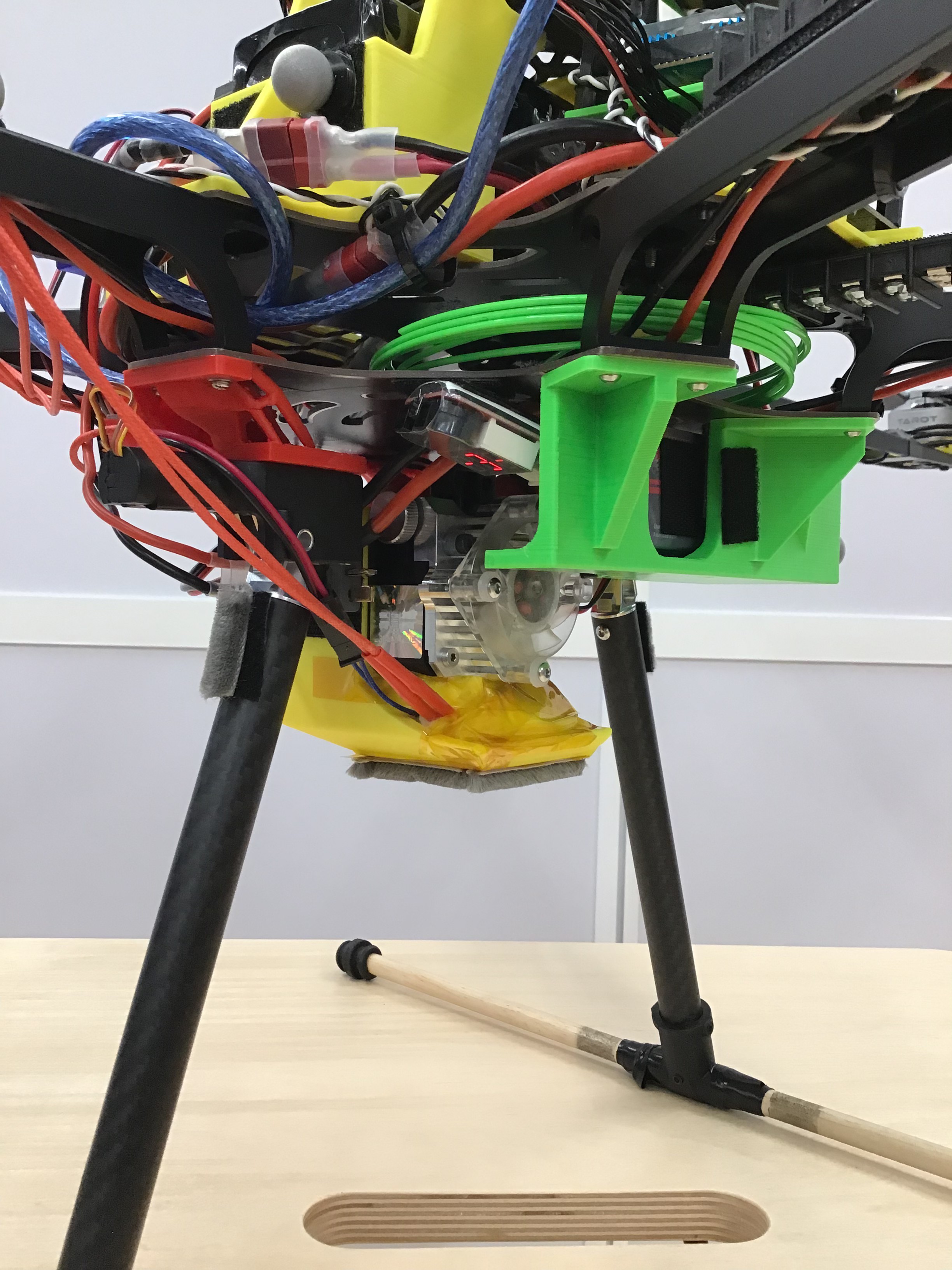} %trim={20cm 10cm 10cm 20cm}
		\label{fig:nozzleGuardYellow}
	}
	\subfigure[Nozzle guard from below]
	{
		\includegraphics[height=0.3\linewidth,  trim={42cm 5cm 12cm 35cm}, clip=true]{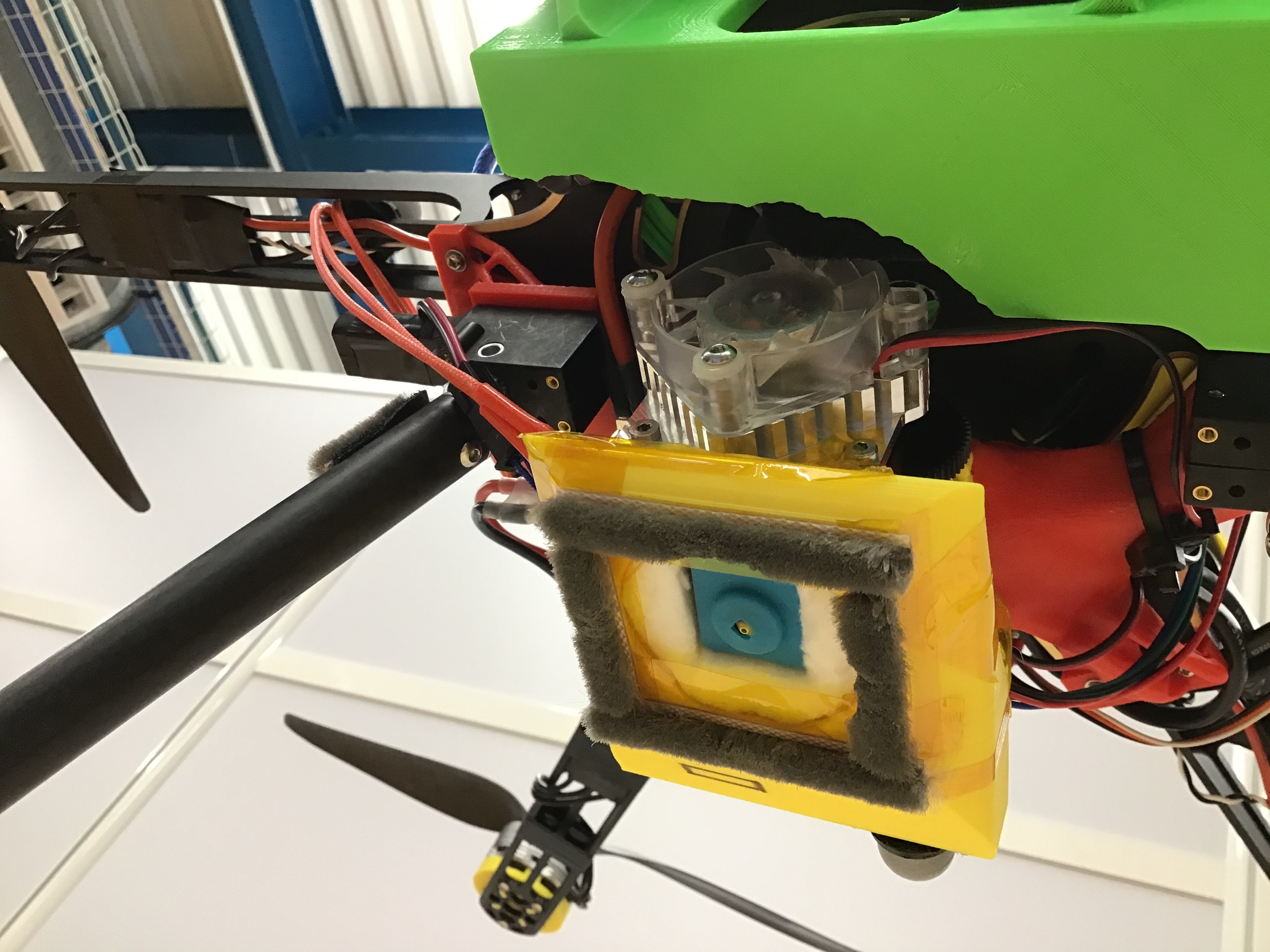}
		\label{fig:nozzleGuardHole}
	}
	\caption{Close-up of the 3D printing apparatus and the nozzle guard. The hexacopter touches the surface with the nozzle guard and uses the nozzle guard to slide across the build surface.}
	\label{fig:apparatus}
\end{figure}

The above components are already sufficient for 3D printing. However, given the design objectives, we cannot guarantee that the nozzle will not touch the build surface during printing or that the nozzle's distance to the surface will be constant, even with advanced control algorithms. Thus, we design a nozzle guard that surrounds the extruder nozzle (yellow part in Fig.~\ref{fig:nozzleGuardYellow}). The hexacopter uses the nozzle guard to touch the build surface and slide across the surface. While sliding, the nozzle prints through a hole in the nozzle guard as shown in Fig.~\ref{fig:nozzleGuardHole}. The distance will be nearly constant as long as the hexacopter does not deviate much from the upright position during printing. 

In addition to guaranteeing a nearly-constant distance to the build surface, the nozzle guard also significantly impacts the goal of printing with high positional accuracy. By creating friction between the nozzle guard and the build surface, the hexacopter is more robust to disturbances from ground effects and general inaccuracies in the position control. The more friction the hexacopter has with the surface, the fewer the disturbances that need to be compensated for by the motors. However, if there is too much friction, the hexacopter needs to use more effort to slide during printing. To prevent hard surface contact, we install thin weather striping underneath the nozzle guard.

The nozzle guard also accommodates ceramic fiber insulation that insulates the hotend from the surrounding air flow. Insulating the hotend makes the heating of the hotend more efficient, preserving battery power. Furthermore, the insulation protects the nozzle guard from melting as the nozzle guard is out of the same PLA material the 3D printing hexacopter uses to print.

\subsection{Balancing the hexacopter}

It is well known that for position accuracy and stability purposes the horizontal location of a drone's center of mass should coincide with its geometric center. We design the testbed with this requirement in mind by distributing the weight evenly across the platform. However, since the 3D printing hexacopter uses parts with different masses, the center of mass is still off from the geometric center. To balance the hexacopter, we iteratively adjust counterweights along the hexacopter's arms until the horizontal location of the center of mass aligns with the geometric center. 

\section{CONTROL STRATEGY}\label{sec4}

Before we can test the designed testbed, we need to develop a control strategy for sliding across the build surface. To perform initial print tests, we focus on keeping the control strategy simple and easily tunable. This approach will allow us to observe the behavior of the 3D printing hexacopter and modify the controller during the experiments.

As mentioned in Section~\ref{sec3}, the hexacopter is designed to slide across the surface. Since the build surface acts as a support in the vertical direction, we can separate the controller design for the horizontal and vertical directions. We first discuss the horizontal position controller and then follow with a thrust controller for the vertical direction. These controllers are activated once the hexacopter reaches the build surface. 

\subsection{Horizontal position controller}

To control the horizontal position of the 3D printing hexacopter, we use a simple PID position controller that outputs roll and pitch commands to the PX4 flight stack. The yaw is always zero. PX4 then converts the attitude commands to motor outputs through its internal control structure~\cite{px4}. The pitch angle at time step $k$ is calculated with respect to the fixed world frame of the motion capture system by 
\begin{align}
\theta[k] =
\begin{cases}
\alpha_{lim} + \beta_\theta & \ \text{if }\ \theta^\prime[k] \geq \alpha_{lim} + \beta_\theta \\
\beta_\theta -\alpha_{lim}  & \ \text{if }\ \theta^\prime[k] \leq \beta_\theta -\alpha_{lim} \\
\theta^\prime[k] & \ \text{otherwise,}
\end{cases} 
\end{align}
where $\alpha_{lim}$ is an attitude limit, and $\beta_\theta$ is a tuned parameter that accounts for errors in the attitude estimation. $\theta^\prime[k]$ is calculated by the modified PID controller
\begin{gather}
\theta^\prime[k] = K_{p,\theta} \ e_x[k] + K_{i, \theta} \ i_x[k] + K_{d, \theta} \ d_x[k] + \beta_\theta,
\end{gather}
with the individual terms given by
\begin{gather}
e_x[k] = x_{d}[k] - x_{m}[k] \label{eq:error} \\
i_x[k] = i_x[k-1] + \Delta t \ e_x[k] \\
d_x[k] = \frac{e_x[k] - e_x[k-1]}{\Delta t}.
\end{gather}
Here, $\Delta t$ is the time difference between updates of the position controller, and $K_{p,\theta}$, $K_{i, \theta}$, and $K_{d, \theta}$ are the proportional, integral, and derivative gains, respectively. $x_d[k]$ and $x_m[k]$ are the desired setpoint and the measured position in the x-axis of the world frame. For the experiments of this paper, the world frame uses the east-north-up convention. To calculate the roll angle, we use the same equations but with a flipped sign for the calculated angle, the measured position $y_{m}[k]$ in the y-direction, and the desired setpoint $y_{d}[k]$.

Compared to a regular PID controller, the above controller features several modifications. First, $\alpha_{lim}$ ensures that the hexacopter stays close to the upright position. Staying close to this equilibrium position keeps the distance between the nozzle and the build surface nearly constant. $\alpha_{lim}$ also prevents any drastic maneuvers that could cause a crash during printing. For the experiments of this paper, $\alpha_{lim}$ has a value of five degrees. Second, we use ramping setpoints to generate smooth printing motions and prevent the position controller from saturating. At each time step, the current setpoint is updated towards the commanded printing destination based on the allowed sliding speed during printing. 

Finally, we use $\beta_\phi$ and $\beta_\theta$ to account for a constant offset in the attitude estimation from the actual attitude. The hexacopter might be in its upright position, but the flight controller may estimate the roll and pitch to be somewhere around $\pm 1$ degrees. This offset stems from the IMU calibration of the hexacopter and can usually be ignored as the offset is relatively small compared to the commanded attitudes in free flight. However, since we are using small angles to control the sliding motion during 3D printing, this offset can significantly impact the horizontal position controller. 

Though it is possible to reduce this offset by iteratively tuning the parameters for PX4, the calibration process is tedious. The calibration requires iterations of taking off, observing the hexacopter's flight behavior, landing, and recalibrating the parameters with slight hardware modifications. Yet, these iterations may still not eliminate the offset. Thus, we use $\beta_\phi$ and $\beta_\theta$ to account for this offset. 

Finding $\beta_\phi$ and $\beta_\theta$ is straightforward; we let the hexacopter hover a few meters above the ground to eliminate any ground effects and then command the hexacopter to have zero roll, pitch, and yaw. If there is an offset, the hexacopter will start moving towards the corresponding direction. We change the commanded angles iteratively until the hexacopter no longer drifts from its starting position. These angles are then used as bias for the horizontal position controller. 

\subsection{Thrust controller}

An important assumption for the horizontal position controller is that the friction between the nozzle guard and the build surface stays roughly constant. Since the PID gains depend on this friction, a drastic change in friction can result in instabilities of the hexacopter or the hexacopter not moving at all. To keep constant friction, one could send a constant thrust command to PX4, which will result in constant RPMs of the motors. However, due to the voltage drop of the battery, the RPMs would continuously decrease, increasing the friction as a result. 

\begin{figure}
	\centering
	\begin{tikzpicture}[scale=0.9]
	\begin{axis}[
	height=4.5cm,
	width=0.9\linewidth,
	axis lines = left,
	xlabel = {$ Voltage \ Percentage \ [\%]$},
	ylabel = {$Commanded \ Thrust \ [\%]$},
	ymin=43.5,
	ymax=49
	]
	%Below the red parabola is defined
	\addplot [
	domain=35:100, 
	samples=100, 
	color=blue,
	]
	{-60 * (x/100 - 0.62)^3 + 47};
	% %{-60 * (x/100 - 0.62)^3 + 47}; %{min(0.48,max(0.43,-0.926 * (x - 0.65)^3 + 0.455))};
	\end{axis}
	\end{tikzpicture}
	\caption{Cubic function used for updating the commanded thrust during printing. For the experiments of this paper, we replace the battery at roughly 40 percent remaining voltage.}
	\label{fig:thrust}
\end{figure}
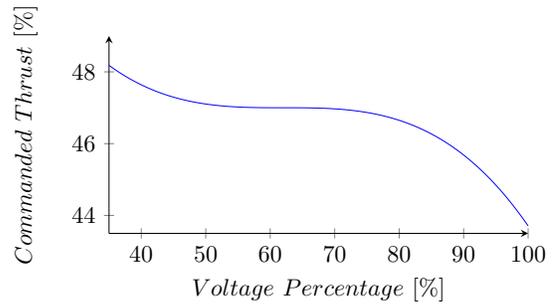

To counter this voltage drop during printing, we update the commanded thrust to PX4 based on the remaining voltage percentage. The remaining voltage percentage is calculated by \[V_{remaining} = 100 \cdot \frac{V_{measured} - V_{min}}{V_{max} - V_{min}}, \] where $V_{measured}$ is the measured voltage, $V_{max}$ is the voltage of the fully-charged battery, and $V_{min}$ is the minimum voltage still considered safe for the battery. The voltage of a lithium polymer battery usually decreases quickly at the beginning of a discharging cycle and then continues to decrease at a slower pace until the voltage is beyond the usual operating range. To roughly account for this behavior, we update the commanded thrust based on a cubic function plotted in Fig.~\ref{fig:thrust}. The commanded thrust is significantly reduced at the start when the battery is fully charged and then quickly reaches a point where the thrust stays relatively constant.

\section{EXPERIMENTAL EVALUATION}\label{sec5}

In this section, we experimentally evaluate the 3D printing hexacopter. We first discuss the printing procedure and then present the experimental results. We conclude this section with a discussion of the results and the 3D printing hexacopter concept.

\subsection{Printing procedure}

To demonstrate the hexacopter's printing capabilities, we print simple contours on the surface of a box. The box's designated printing area is covered with blue printer tape to increase adhesion with the PLA material. At the start of each printing experiment, the hexacopter automatically flies to the box and descends until the nozzle guard touches the build surface. Once the hexacopter makes contact, we switch from the PX4 position controller to the horizontal position controller from Section~\ref{sec4}. To traverse the desired path, we sequentially update the position controller's setpoints to the contours' corner coordinates. While the hexacopter moves across the surface, the 3D printing apparatus continuously deposits material onto the build surface. 

\subsection{Results}

\begin{figure}
	\centering
	\subfigure[3D printed contour of a square.]
	{
		\includegraphics[width=0.8\linewidth, trim={20cm 10cm 10cm 20cm}, clip=true]{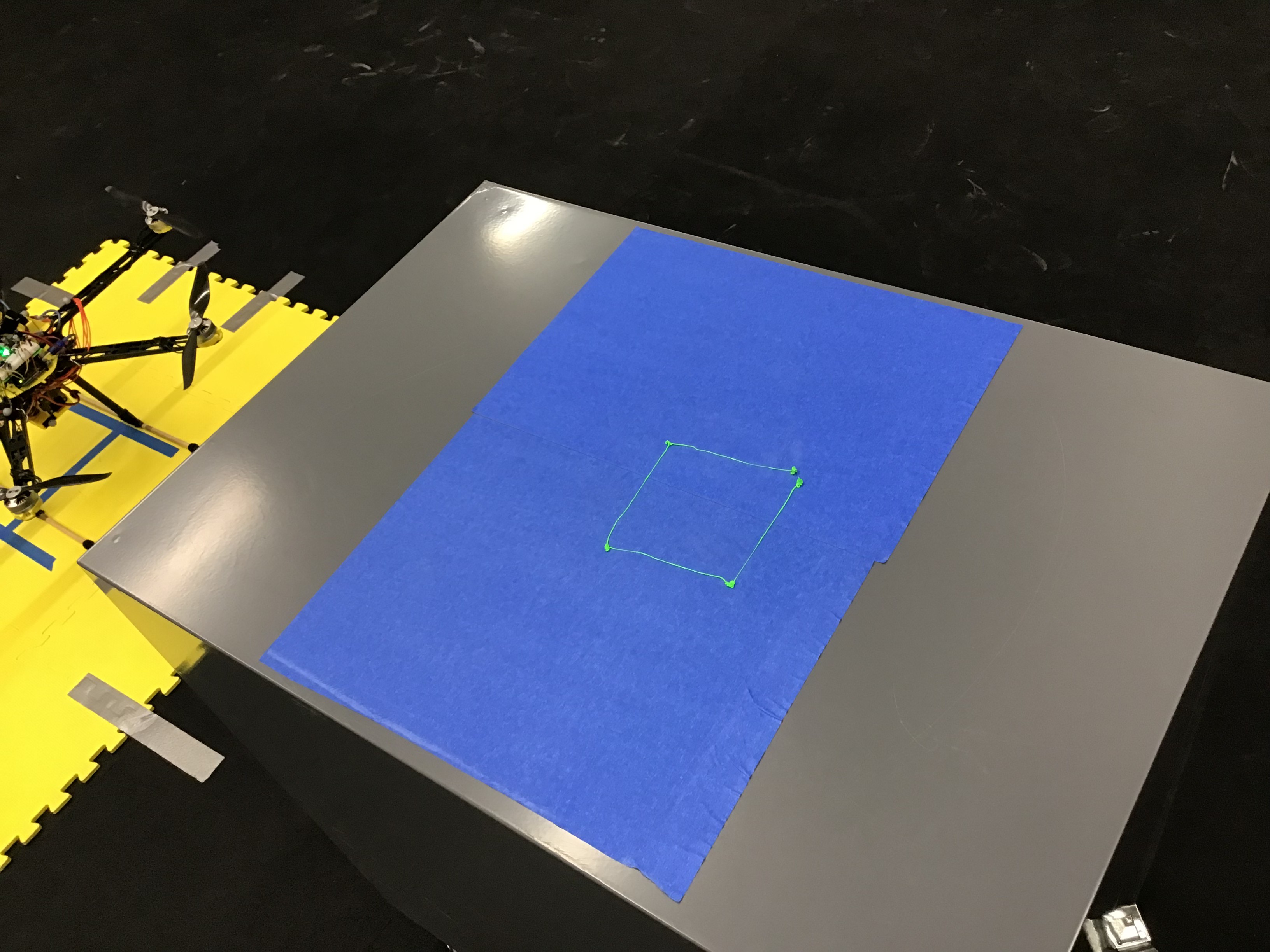} %trim={20cm 10cm 10cm 20cm}
		\label{fig:square}
	}
	\\
	\subfigure[3D printed letters "UT", short for University of Texas.]
	{
		\includegraphics[width=0.8\linewidth, trim={25cm 23cm 15cm 53cm}, clip=true]{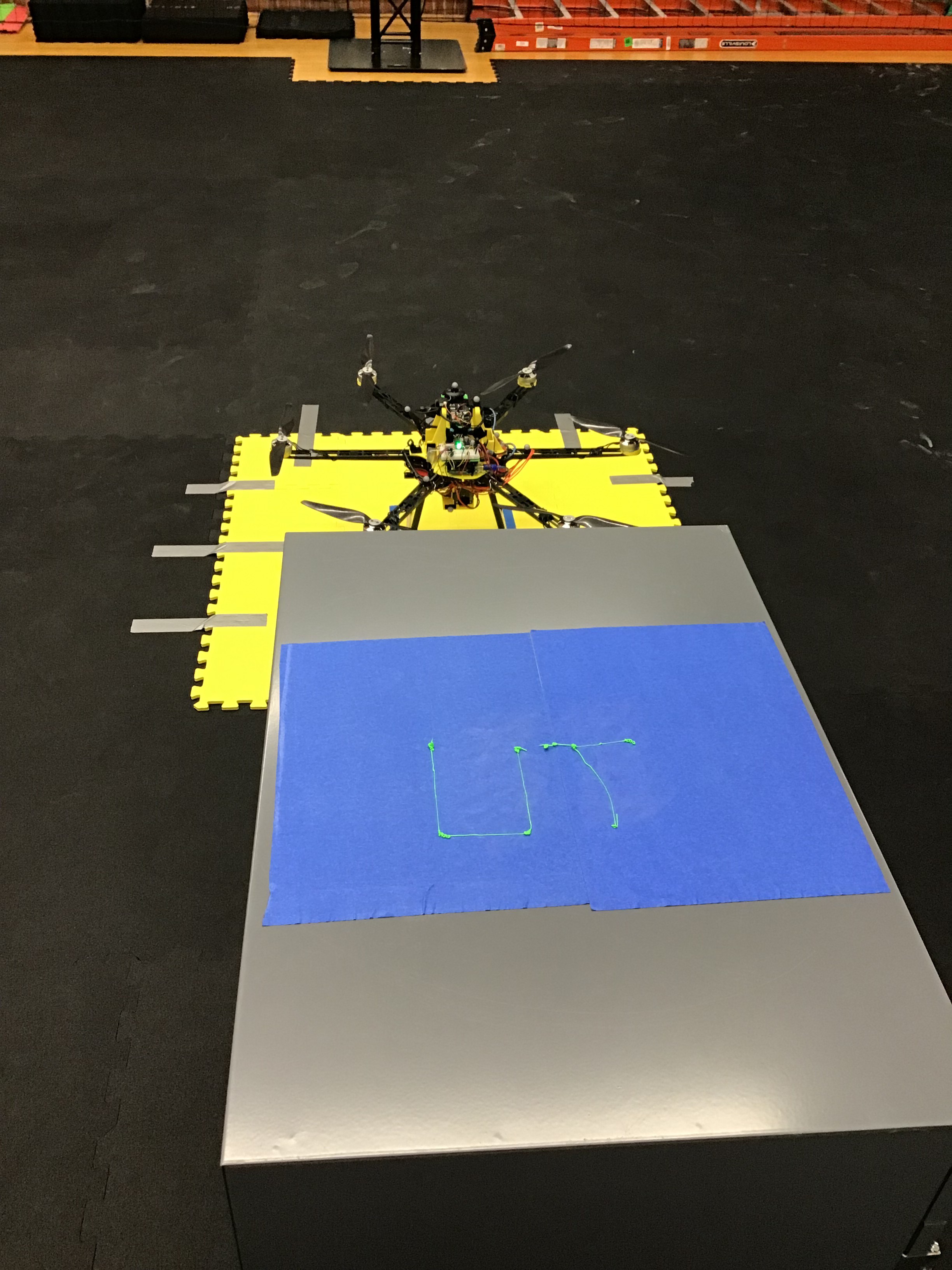}
		\label{fig:UT}
	}
	\caption{Simple contours printed by the 3D printing hexacopter on blue printer tape.}
	\label{fig:print_results}
\end{figure}

We demonstrate the hexacopter's capability to successfully 3D print by presenting two examples of the printed contours in Fig.~\ref{fig:print_results}. The first contour is a square with side lengths of \mbox{10 cm}. The second contour exhibits the letters "UT". The printed contours show that the deposited PLA materials stick to the surface and can withstand the turbulent air from the hexacopter. We provide a video showing the successful print of the square contour.~\footnote{https://youtu.be/tEooDpE2TyE}

Based on the position measurements from the motion capture system, the hexacopter's overall position accuracy is approximated to be $\pm 1$ cm in the horizontal directions. This tolerance is especially noticeable when the hexacopter starts the sliding motion or reaches the desired corner setpoint. However, once the hexacopter starts sliding, the accuracy appears to be significantly better.

From visual inspections of the printed contours, we notice that the edges feature relatively uniform material deposition and minimal variation within the straight lines. These results indicate that the hexacopter successfully maintains a constant and tight distance between the nozzle and the build surface, largely due to the nozzle guard. From running multiple consecutive print tests, the nozzle guard also proves to properly shield the hotend, allowing it to maintain a constant temperature of $250~\degree C$ while protecting the nozzle from ground contact.

The corners of the printed contours exhibit unwanted clumps of PLA material. These clumps are a result of the hexacopter waiting for a new corner setpoint while the 3D printing apparatus continued to deposit material. A consequence of these clumps is the curved vertical line in the "T" letter of the second printed contour. As the hexacopter moved across the surface, the top clump of the vertical line was dragged to the left by the weather stripping underneath the nozzle guard.

\subsection{Discussion}

The initial printing results demonstrate the feasibility of combining 3D printing with drones as the hexacopter testbed can print simple contours on the build surface. However, since the 3D printing hexacopter is a novel testbed, several limitations still need to be addressed to improve the current printing capabilities. First, a trajectory tracking algorithm for simultaneously controlling the hexacopter's movement and printing is required to eliminate the observed PLA clumps. Second, the current position accuracy needs to be further quantified. Based on this quantification, the accuracy can be improved by developing advanced control algorithms customized for the 3D printing process. Finally, the complex interactions of the nozzle guard and the ground need to be investigated further.

In addition to these immediate improvements, general questions about the concept of a 3D printing drone may arise. For example, a drone can only stay in the air for a limited amount of time and carry a limited amount of 3D printing material. However, a tethered setup for power and material supply are promising options to ensure continuous printing. Though a tethered setup may limit a drone's flexibility, the tethered connection can be easily modified to fit the desired end-application. This modification allows the drone to maintain its flexibility and advantage over robotic arms or other 3D printing machines for large-scale objects. Another question may address the hexacopter's dependency on the nozzle guard. When building objects in the vertical direction, the nozzle guard will quickly lose contact with the ground. However, a modified nozzle guard design, similar to wall-climbing robots, may maintain the nozzle guard's functionality by using the previously printed object or another close-by object as support. Furthermore, dependencies on the nozzle guard may be reduced by advanced control algorithms customized for the 3D printing process.

Overall, we demonstrate the feasibility of combining 3D printing with drones by successfully performing initial FDM print tests. In doing so, we lay the groundwork to overcome limitations in size and location for 3D printed parts and structures.

\section{CONCLUSIONS}\label{sec6}

Parts built by 3D printing robots or other 3D printing machines are usually limited in size or location by the 3D printer. To overcome these limitations, we built a hexacopter testbed that can deposit material on a build surface via a 3D printing process called fused deposition modeling. With the goal of printing with high positional accuracy, we discussed the hexacopter's design and developed a simple control strategy for 3D printing while flying. We successfully performed initial print tests by printing simple contours on a build surface. In doing so, we demonstrated the feasibility of our approach. Future work will focus on the investigation of advanced control algorithms to further improve the hexacopter's printing accuracy.

\bibliographystyle{IEEEtran}
\bibliography{references}

\end{document}